\documentclass{article}
\usepackage{spconf,amsmath,graphicx}

\usepackage{amsfonts,amssymb}
\usepackage{booktabs}
\usepackage{multirow}
\usepackage{subfigure}

\title{BPDO:Boundary Points Dynamic Optimization for Arbitrary Shape Scene Text Detection}
\name{Jinzhi Zheng$^{1,2}$ \qquad Libo Zhang$^{1,2} \sthanks{Corresponding author: libo@iscas.ac.cn}$ \qquad Yanjun Wu$^{1}$  \qquad  Chen Zhao$^{1}$}
\address{
$^{1}$Institute of Software Chinese Academy of Sciences, Beijing, China \\
$^{2}$ University of Chinese Academy of Sciences, Beijing, China\\
}
\begin{document}
%
\maketitle
\begin{abstract}
Arbitrary shape scene text detection is of great importance in scene understanding tasks. Due to the complexity and diversity of text in natural scenes, existing scene text algorithms have limited accuracy for detecting arbitrary shape text. In this paper, we propose a novel arbitrary shape scene text detector through boundary points dynamic optimization(BPDO). The proposed model is designed with a text aware module (TAM) and a boundary point dynamic optimization module (DOM). Specifically, the model designs a text aware module based on segmentation to obtain boundary points describing the central region of the text by extracting a priori information about the text region. Then, based on the idea of deformable attention, it proposes a dynamic optimization model for boundary points, which gradually optimizes the exact position of the boundary points based on the information of the adjacent region of each boundary point. Experiments on CTW-1500, Total-Text, and MSRA-TD500 datasets show that the model proposed in this paper achieves a performance that is better than or comparable to the state-of-the-art algorithm, proving the effectiveness of the model.

\end{abstract}
\section{Introduction}
\label{sec:intro}

Scene text detection, as fundamental research in visual scene understanding, has long received widespread attention due to its common applications in areas such as autonomous driving, blind assistance, intelligent shopping, visual translation, etc\cite{14,44}. As a key step in scene understanding, the purpose of scene text detection is to detect text in the scene image and locate the position of the text. Thanks to the rapid development of artificial intelligence\cite{42,43}, algorithms \cite{10,11,19} based on neural networks in artificial intelligence have gradually been proposed, making great progress in scene text detection algorithms. However, existing scene text detection algorithms are still challenging, mainly due to the wide range of complexity and diversity of scene text. 

\begin{figure}[t]
\centering
\includegraphics[width=0.4\textwidth]{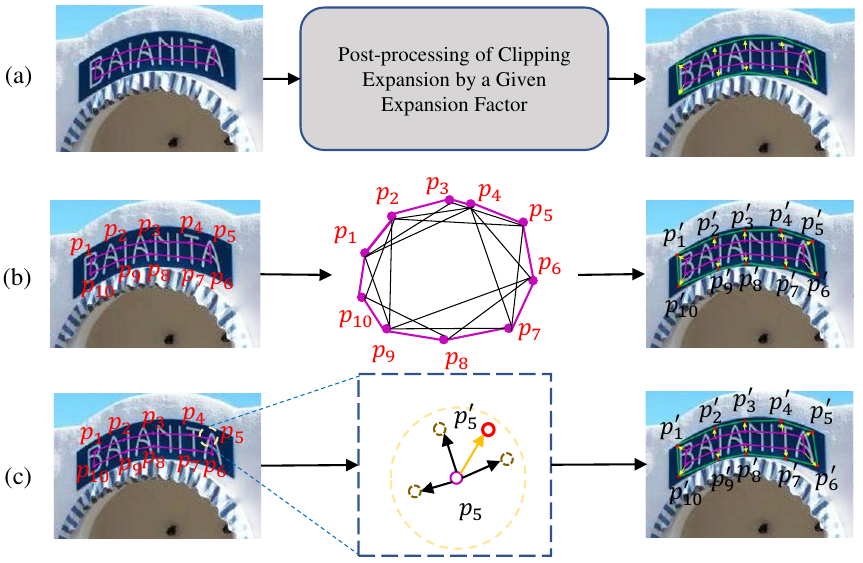}
\caption{ Comparison of BPDO with other segmentation-based scene text algorithms. (a) Segmentation-based algorithm for scene text detection. (b) Scene Text Segmentation Algorithm for boundary points  optimisation. (c) BPDO. Each boundary point is progressively optimized for the text region through neighborhood information.}
\label{fig: teaser}
\end{figure}

\begin{figure*}[t]
\centering
\includegraphics[width=0.94\textwidth]
{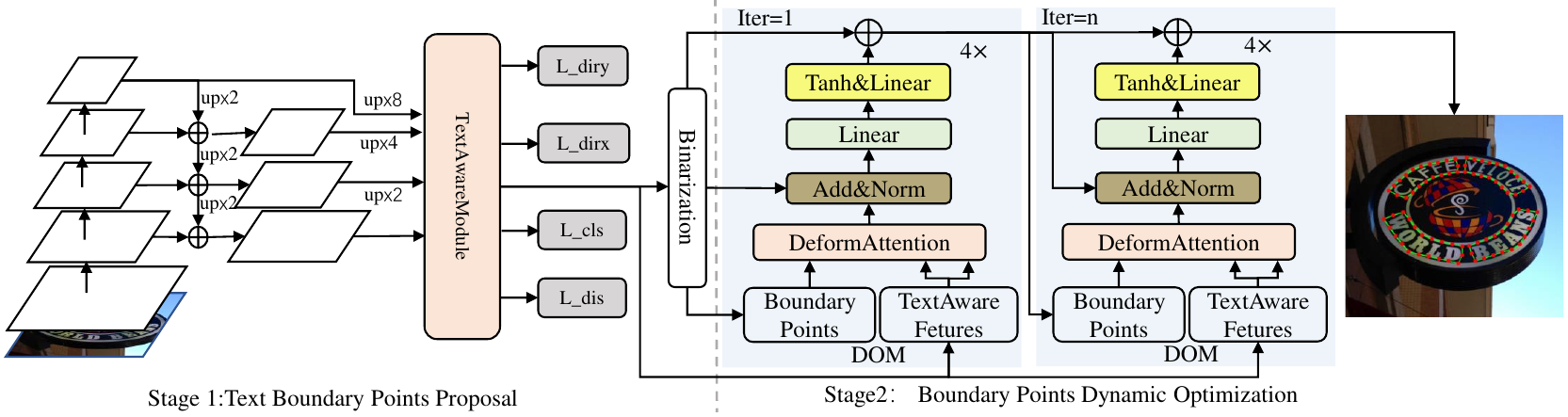}
\caption{The overall architecture of the Boundary Points Dynamic Optimization. $L\_dis$ and $L\_cls$ represent distance map and segmentation map. $L\_dirx$ and $L\_diry$ represent directional map in the x and y directions, respectively.}
\label{Figure2.}
\end{figure*}

Segmentation-based algorithms \cite{10,11,14,15} with pixel-level accuracy have gained widespread attention in text detection of arbitrary shapes.
To avoid segmenting adjacent text into a connected region, which can cause text detection failure, these methods usually perform segmentation on the central region of the text and then expand the detected central region to the complete text region according to a certain threshold, as shown in Fig.\ref{fig: teaser}(a). This type of method is able to satisfy the detection of arbitrary shape text but is sensitive to noise. To reduce the effect of noise on the detection results and further improve the detection accuracy of scene text, some algorithms\cite{19,21,22} detect scene text by predicting or generating boundary points, as shown in Fig.\ref{fig: teaser}(b). This type of method is divided into two steps: The first step, based on the visual features, generates the coarse boundary points. In the second step, based on the coarse boundary point, obtain the offset of the coarse boundary point through LSTM or transformer, and modify the boundary points to obtain the border of the complete text regions. However, as shown in Fig.\ref{fig: teaser}(b), such algorithms based on boundary points modeling usually perform position optimization or grouping between boundary points by LSTM, GCN, or Transformer. The dynamic optimization process of boundary points ignores the feature relationship between the boundary point and its adjacent spatial pixels, thus limiting the accuracy of text detection.

We argue that in the optimization process of boundary points, the information in the space adjacent to the boundary point is more important than that brought by other farther boundary points. Therefore, in order to further improve the detection accuracy of arbitrary shape scene text, inspired by the idea of particle filter \cite{25} and based on the implementation of the deformable attention \cite{24}, this paper proposes a boundary points dynamic optimization model for arbitrary shape scene text detection. Meanwhile, in order to make the extracted visual features better adapted to the segmentation of text instances at different scales, we also designed a text aware module. The main contributions of this paper are as follows:(1) A text aware module is designed to fuse different scales of visual features from this channel attention level and spatial attention level, respectively. (2) A boundary points dynamic optimization model is proposed to make optimization from the adjacent space based on the current state of the boundary points. (3) Experimental results on the MSRA-TD500, CTW1500, and Total-Text datasets show that the proposed achieves detection performance that is better than or comparable to the state-of-the-art algorithm.

\section{Proposed Method}
\label{sec:Method}
The overall structure of our method is shown in Fig.\ref{Figure2.}. The text detection process is divided into two stages: the text boundary points proposal and the boundary points dynamic optimization. First, we use Resnet50 with FPN and DCN \cite{23} as the backbone to extract visual features. Then, we designed a text-aware module to sense text regions based on the features extracted by the backbone and obtain text aware features under the supervision of a priori information (distance map, direction map, and classification map). Based on the text aware features, the text boundary proposal is generated by a binarized distance map. The text boundary proposal is composed of $N$ points on the text boundary. The second stage: Based on the text aware features, the dynamic optimization module (DOM) optimizes boundary points in an iterative way to obtain accurate text borders and complete text detection.

\begin{figure}[t]
\centering
\includegraphics[width=0.45\textwidth]{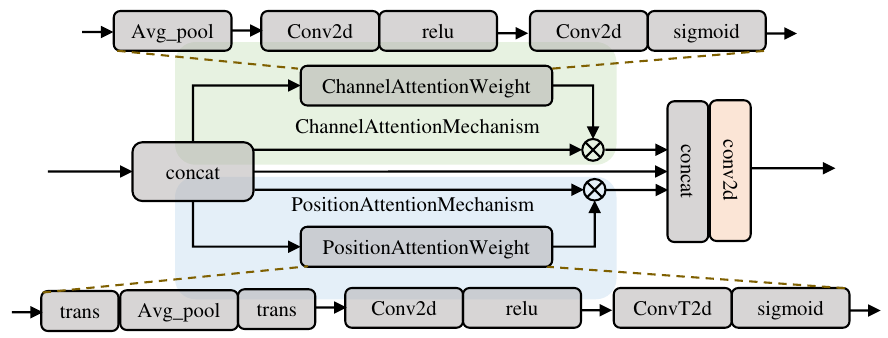}
\caption{ The detailed structure of the text aware module.}
\label{Figure3.}
\end{figure}

\subsection{Text Aware Module}
We used ResNet50 with FPN and DCN \cite{23} as the backbone to extract visual features. Given the scene image $I\in \mathbb{R}^{H\times W \times 3}$, the visual features can be formulated as:
\begin{equation}
\left\{\begin{array}{l}
f_1,f_2,f_3,f_4=R(I) \\
R_v=concat(f_1^\prime,\ f_2^\prime,{\ f}_3^\prime{,f}_4^\prime)
\end{array}
\right.
\end{equation}
Where $C$ is the dimension of the feature channel and $R( )$ represents Resnet50 with FPN and DCN. $f_i^\prime \in \mathbb{R}^{\frac{H}{4}\times\frac{W}{4}\times C}$ is the upsample of $f_i$. The process of text aware features can be formulated as:
\begin{equation}
\left\{\begin{array}{l}
W_c=Channel\_Attention(R_v) \\
W_p=Position\_Attention(R_v)\\
F_{tam}=Conv(Concat(R_v\ast W_c,R_v\ast W_p,R_v)
\end{array}
\right.
\end{equation}
where $ W_c$ and $W_p$ are the channel attention weights and position attention weights, respectively.
$Channel\_Attention()$, and $Position\_Attention()$ are the solving process for the corresponding attention, respectively. As shown in Fig.\ref{Figure3.}, $Channel\_Attention()$ consists of an average pooling layer, a convolution layer with $relu$, and a convolution layer with $sigmoid$. $Position\_Attention()$ needs to be transposed before and after the average pooling layer, and then processed by convolution with $relu$ and deconvolution with $sigmoid$. Prior information (distance map, direction map, classification map) of the text is predicted based on text aware features. Boundary points \cite{19,21} of the scene text are proposed based on the classification map and direction maps as text boundary proposals. In this paper, 20 boundary points are sampled on the boundary of each text instance.

\subsection{Dynamic Optimization Module}
Inspired by the Deformable Attention \cite{24} and Particle Filter \cite{25}, we designed a dynamic optimization module (DOM) for the boundary points, which optimizes the boundary points in an iterative manner, as shown in Fig.\ref{Figure2.}. The idea of dynamic optimization based on the position of boundary points and neighborhood information is to randomly sample several points (3 in this paper) in the current neighborhood space of the boundary points, and confirm the movement information of the boundary points by querying and coding the sampled point, as shown in Fig.\ref{fig: teaser}.

As shown in Fig.\ref{Figure2.}, the dynamic optimization module (DOM) mainly consists of deformable attention \cite{24} to achieve the optimization of boundary points in the neighborhood space. Deformable attention can be formulated as:
\begin{equation}
\left\{\begin{array}{l}
V_m^n=W_m^{'}(F_{tam}(P+\Delta _m^n)) \\
W_{mn}^{qk}=W_m^n(F_{tam}(P)) \\
F_{da}=\sum_{m=1}^{M}W_m[\sum_{n=1}^{N}W_{mn}^{qk}*V_m^n)]
\end{array}
\right.
\end{equation}
Where $M$ and $N$ represent the number of attention heads (set as 8 in this paper) and the number of random sampling points per optimization (set as 3 in this paper), $W_m^{'}()$ and $W_m^n()$ represent the fully connected layer linear transformation. $F_{tam}(P+\Delta _m^n)$, $F_{tam}(P)$ represent the values of $F_{tam}$ at the $P+\Delta _m^n$ and $P$ positions, $P$ and $\Delta _m^n$ represent the current boundary point position coordinates and their offsets, $\Delta _m^n$ is obtained by the linear transformation of $F_{tam}(P)$ by the fully connected layer.

\subsection{Multi-objective loss function}
In the process of model training, similar to \cite{19,21}, prior information is required for supervised learning, so the same multi-objective loss function is used:
\begin{equation}
L\ =\ L_{cls}+\alpha{\times L}_{dis}\ +\ \beta\times L_{dir}\ +\frac{\gamma\times L_{pm}}{1+e^{(i-eps)/eps}}\ 
\end{equation}
Where $L_{cls}$ is the cross entropy loss of text classification map, $L_{dis}$ is the $L_2$ regression loss of distance map, and $L_{dir}$ is the $L_2$ norm loss of distance and angle of direction map. $L_{pm}$ is the matching loss of boundary points \cite{19}. $eps$ is the maximum training epoch and $i$ is the $i-th$ epoch. $\alpha$, $\beta$, and $\gamma$ are balance factors, which are set to 1.0, 3.0, and 0.1 respectively in the following experiment.

\section{Experiment}
\label{sec:Experiment}
\subsection{Datasets}
\label{sec:Datasets}
\noindent {\bfseries SynthText} \cite{26} is a synthetic dataset that generates text by rendering scene text images. The dataset is mainly used to alleviate the problems of difficult sample annotation and insufficient training samples. The dataset contains 800K images. Text instances are annotated at both character and text levels.

\noindent {\bfseries Total-Text}\cite{27} consists of a training set containing 1255 images and a test set containing 300 images. Provide word-level annotations of text instances.

\noindent {\bfseries CTW1500} \cite{28} is a curved text dataset consisting of a training set containing 500 images and a test set of 1000 images. Each scene instance is annotated by 14 points.

\noindent {\bfseries MSRA-TD500} \cite{29} contains multi-directional text, often taken from directional signs, in Chinese, and English. The training set contains 500 images and the test set contains 200 images. The scene text is annotated in lines.

\subsection{Implementation Details}
\label{sec:majhead}
In order to make the comparison as fair as possible, we use the training strategies commonly used by other state-of-the-art algorithms. Similar to \cite{19,21,15,30}, Resnet50 with FPN and DCN is used as a backbone to extract visual features. Since we use TextBPN++ \cite{21} as the baseline and share the backbone with it, we download the trained model from its official website and extract its weights for initialization. The training stage is divided into the pre-train and fine-tune. When training, the image is resized $640 \times 640$. For the pre-training on Synthtext, the epochs are set to 1200, the learning rate is 0.0001, the decay is 0.9 after every 100 epochs, the batch size is 3, and the optimizer is Adam. In the fine-tuning, the batch size is 4, the learning rate is 0.00005, and the optimizer is Adam. The number of border points sampled is 20 and the number of iterations is set to 3.  The algorithm has Python 3.7.13 and Pytorch 1.7.0 implementation. The model was trained and tested using a single NVIDIA TITAN RTX with 24GB.
 
\subsection{Comparisons with State-of-the-Art Methods}
\label{sssec:Comparisons}
\begin{figure}[h]
  \centering
  {\includegraphics[width=0.15\textwidth, height=15mm]{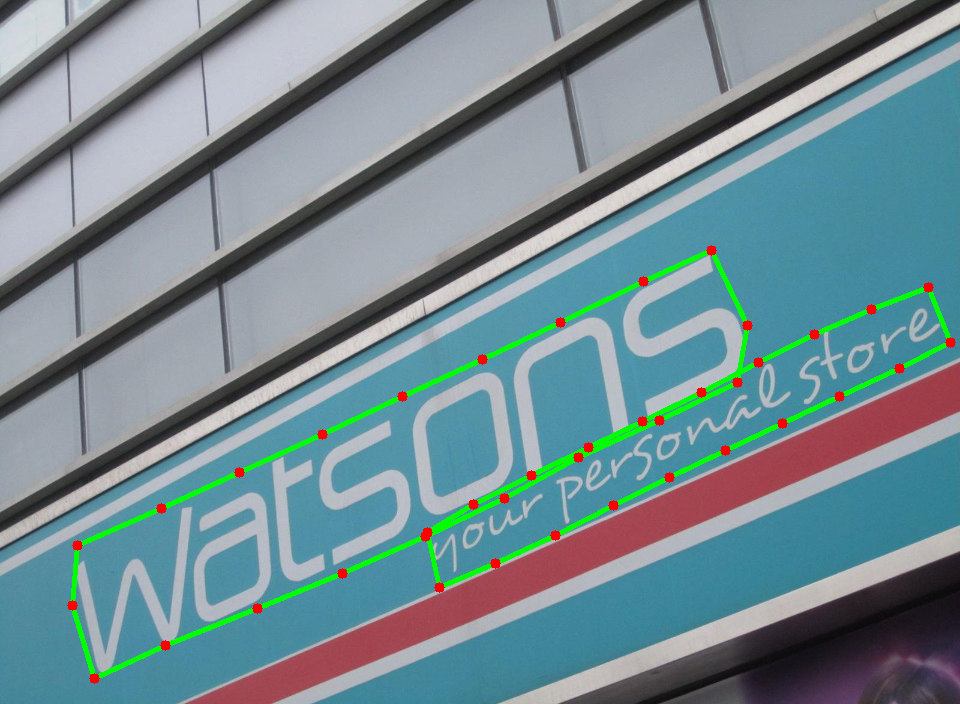}\label{fig:subfig7}} 
  {\includegraphics[width=0.15\textwidth, height=15mm]{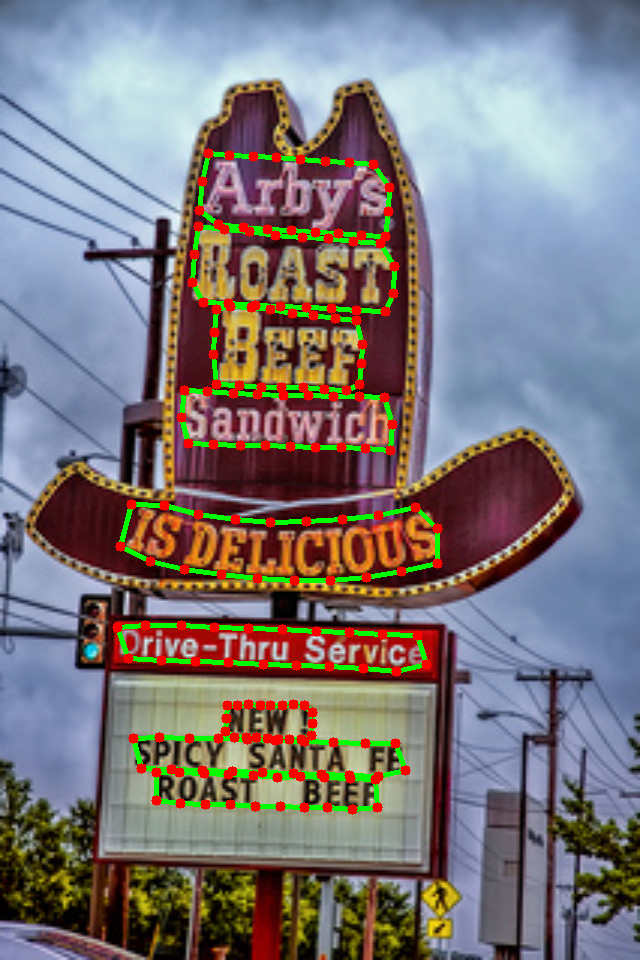}\label{fig:subfig8}}
  {\includegraphics[width=0.15\textwidth, height=15mm]{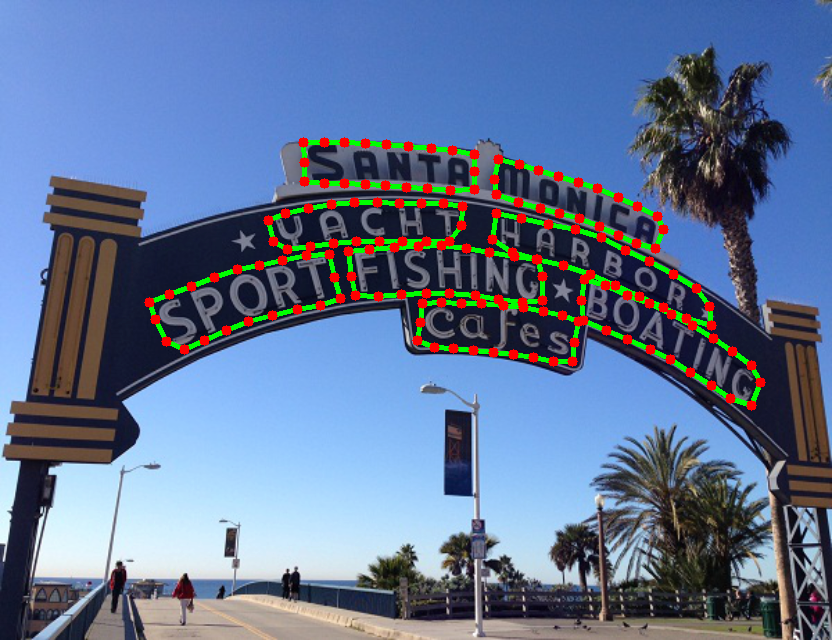}\label{fig:subfig9}}
  
  \subfigure[MSRA-TD500]{\includegraphics[width=0.15\textwidth, height=15mm]{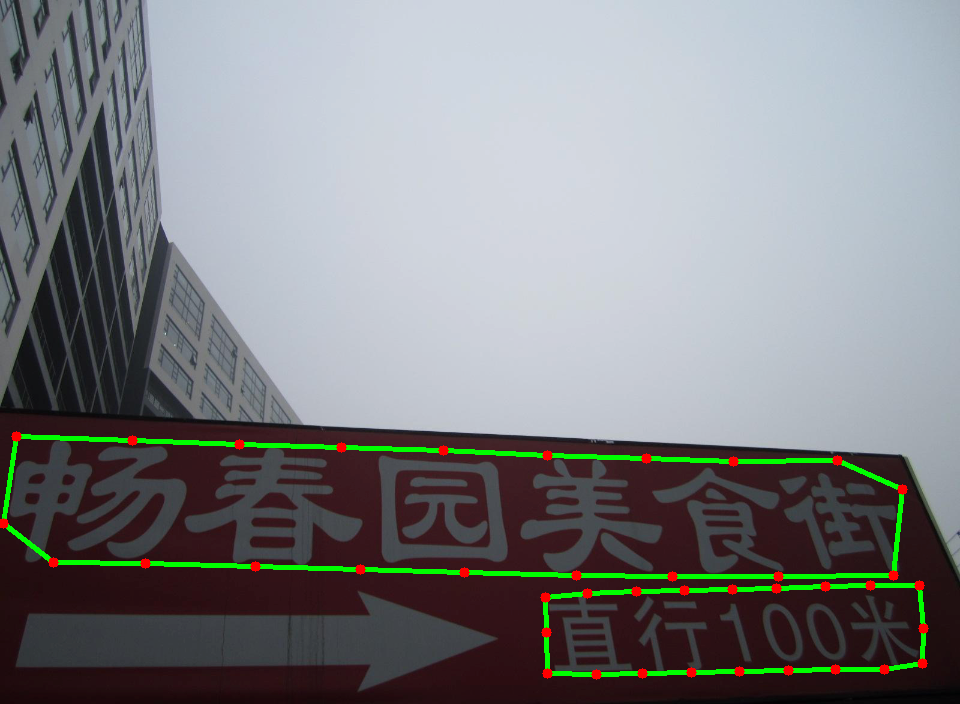}\label{fig:subfig7}} 
  \subfigure[CTW1500]
  {\includegraphics[width=0.15\textwidth, height=15mm]{Figures/Ctw1500/1034_2iter_result.png}\label{fig:subfig8}}
  \subfigure[Total-Text]{\includegraphics[width=0.15\textwidth, height=15mm]{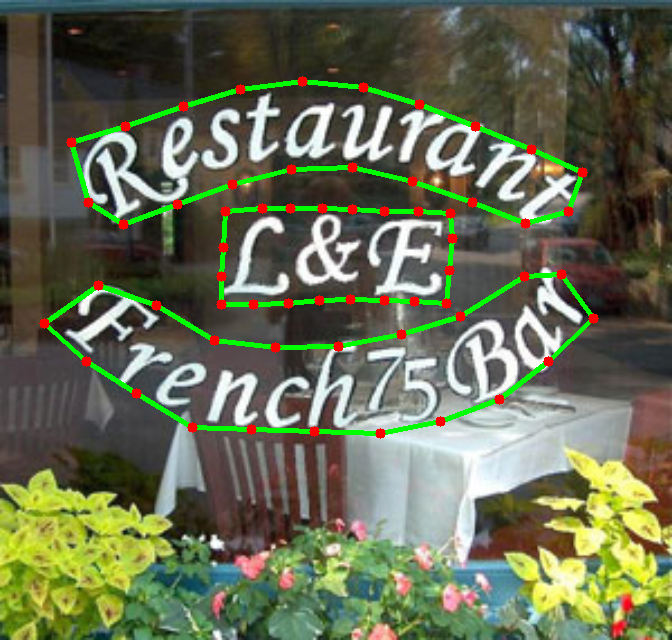}\label{fig:subfig9}}
 
  \caption{Visual results of our algorithm on four datasets. The green irregular polygon is the detected text contours.}
  \label{Figure4.}
\end{figure}

\begin{table*}[h!]
\caption{Text detection accuracy comparison with other methods on MSRA-TD500, CTW1500 and Total-Text. Bold indicates best performance.
}
\setlength{\tabcolsep}{2.8mm}{
\label{Table1.}
\begin{center}
\begin{tabular}{l | l |c c c|c c c | c c c}
\toprule
\multirow{2}{*}{Methods}&\multirow{2}{*}{Papers}& \multicolumn{3}{c|}{MSRA-TD500} &  \multicolumn{3}{c|}{CTW1500} & \multicolumn{3}{c}{Total-Text}  \\ 
\cline{3-11}
\multirow{2}{*}&{}& R(\%) & P(\%) & F(\%) & R(\%) & P(\%) & F(\%) & R(\%) & P(\%) & F(\%)  \\ 
\midrule
LSAE \cite{10} & CVPR'19 & 81.7 & 84.2 & 82.9 & 77.8 & 82.7 & 80.1 & - & - & - \\

CRAFT \cite{31} & CVPR'19 & 78.2 & 88.2 & 82.9 & 81.1 & 86.0 & 83.5 & 79.9 & 87.6 & 83.6\\

PAN \cite{41} & ICCV'19 & 83.8 & 84.4 & 84.1 & 81.2 & 86.4 & 83.7 & 81.0 & 89.3 & 85.0 \\

PSENET  \cite{11} & CVPR'19 & - & - & - & 77.8 & 82.1 & 79. 9 & 75.2 & 84.5 & 79.6 \\
\hline
DRRG \cite{32} &CVPR'20 & 82.3 & 88.1 & 85.1 & 83.0 & 85.9 & 84.5 & 84.9 & 86.5 & 85.7\\

DB \cite{14} &AAAI'20 & 79.2 & 91.5 & 84.9 & 80.2 & 86.9 & 83.4 & 82.5 & 87.1 & 84.7 \\
ContourNet  \cite{33} & CVPR'20 & - & - & - & 84.1 & 83.7 & 83.9 & 83.9 & 86.9 & 85.4 \\
\hline
FCENet \cite{35} & CVPR'21 & - & - & - & 83.4 & 87.6 & 85.5 & 82.7 & 85.1 & 83.9 \\
PCR \cite{36} & CVPR'21 & 83.5 & 90.8 & 87.0 & 82.3 & 87.2 & 84.7 & 82.0 & 88.5 & 85.2 \\
TextBPN \cite{19} & ICCV'21 & 84.54 & 86.62 & 85.57 & 83.60 & 86.45 & 85.00 & 85.19 & 90.67 & 87.85\\
\hline
I3CL \cite{37} & IJCV'22 & - & - & - & 84.6 & 88.4 & 86.5 & 84.2 & 89.8 & 86.9 \\
Bezier \cite{38} & CVPR'22 & 84.2 & 91.4 & 97.9 & 82.4 & 88.1 & 85.2 & 85.7 & 90.7 & 88.1\\
MS-ROCANet \cite{30} & ICASSP'22 &  - & -& -& 83.4 & 88.2 & 85.7 & 83.2 & 85.6 & 84.5 \\
\hline
TCM-DBNet \cite{40} & CVPR'2023& -  & - & 88.8 & - & - & 84.9 & - & - & 85.9 \\

TextBPN++ \cite{21} & TMM'23 & 86.77 & 93.69 & 90.10 &  84.71 & 88.34 & 86.49 &  {\bfseries 87.93} & {\bfseries 92.44} & {\bfseries 90.13}\\
DB++ \cite{15} & TPAMI'23 & 83.3 & 91.5 & 87.2 & 82.8 & 87.9 & 85.3 & 83.2 & 88.9 & 86.0\\
TextPMs \cite{39} & TPAMI'23 & 86.94 & 91.01 & 88.93 & 83.83 & 87.75 & 85.75 & 87.67 & 89.95 & 88.79\\
\hline
BPDO & ours & {\bfseries 88.48} & {\bfseries 94.66} & {\bfseries 91.47} & {\bfseries 84.78} & {\bfseries 88.41} & {\bfseries 86.56} & 84.17 &  90.45 & 87.2 \\
\bottomrule
\end{tabular}
\end{center}
}
\end{table*}
We compare the method proposed in this paper with the current state-of-the-art algorithm as shown in Table \ref{Table1.}. Our algorithm achieves better performance than the state-of-the-art algorithms on the MSRA-TD500. Comparable performance to the state-of-the-art algorithm is achieved on CTW1500 and Total-Text.
For qualitative analysis, we visualized the detection results of the algorithm on three datasets, as shown in Fig.\ref{Figure4.}. It can be seen that our algorithm can correctly detect scene text on all three data sets.

\subsection{Ablation Study}
\label{ssec:Ablation}

\begin{table}[h]
\caption{Ablation study of each module on MSRA-TD500. TAM and DOM represent the text aware module and dynamic optimization module, respectively.}
\begin{tabular}{ c c c c c c c}
\toprule
Method & TAM & DOM & P(\%) & R(\%) & F(\%) \\ 
\midrule
 Baseline & $\times$ &  $\times$ & 93.69 & 86.77  & 90.10\\ 
 Baseline & $\surd$ & $ \times$ & 93.91 & 87.45 & 90.56  \\
 Baseline & $\surd$ & $\surd$ & {\bfseries 94.66} & {\bfseries 88.48} & {\bfseries 91.47} \\
\bottomrule
\end{tabular}
\label{Table2}
\end{table}

In order to verify the effectiveness of each module, we conducted ablation studies on MSRA-TD500, as shown in Table \ref{Table2}. It can be seen that using DOM improves the P, R, and F of baseline by 0.22\%, 0.68\%, and 0.46\%, respectively. Combined use of DOM and TAM improves P, R, and F of baseline by 0.97\%, 1.71\%, and 1.37\%, respectively. This proves that both DOM and TAM designed in this paper are effective.

\section{conclusion}
In this paper, we propose a novel scene text detection network named Boundary Points Dynamic Optimization (BPDO). In this network, we design a text aware module based on position attention and channel attention, as well as a boundary points dynamic optimization module that optimizes boundary points based on boundary point neighborhood information. Unlike the previous state-of-the-art algorithms which optimizes border points according to the relationship between boundary points, the dynamic optimization module proposed in this paper performs self-optimization based on the position and neighborhood information of boundary points. Experiments show that the proposed algorithm achieves better performance than state-of-the-art algorithms on the MSRA-TD500, and has comparable performance with state-of-the-art algorithms on CTW1500 and Total-text. This paper provides a new boundary points optimization idea for scene text detection. In future work, we will further study the end-to-end scene text spotting and recognition algorithm based on the current work.

\bibliographystyle{IEEEbib}

\end{document}